\documentclass[10pt,twocolumn,letterpaper]{article}

\usepackage{cvpr}              

\definecolor{cvprblue}{rgb}{0.21,0.49,0.74}
\usepackage[pagebackref,breaklinks,colorlinks,allcolors=cvprblue]{hyperref}

\def\paperID{00006} 
\def\confName{CVPR}
\def\confYear{2025}

\title{An LLM-enabled Multi-Agent Autonomous Mechatronics Design Framework}

\author{Zeyu Wang\textsuperscript{1}, Frank P.-W. Lo\textsuperscript{1}, Qian Chen\textsuperscript{1}, Yongqi Zhang\textsuperscript{1}, Chen Lin\textsuperscript{1}, Xu Chen\textsuperscript{1}, Zhenhua Yu\textsuperscript{2},\\ Alexander J. Thompson\textsuperscript{1},  Eric M. Yeatman\textsuperscript{1*}, Benny P. L. Lo\textsuperscript{1*}
\\
\textsuperscript{1}Imperial College London, London, UK\\\textsuperscript{2}Univeristy of Aberdeen, Aberdeen, UK\\
{\tt\small e.yeatman@imperial.ac.uk, benny.lo@imperial.ac.uk}
\and
}

\usepackage{bm}
\usepackage{amsmath}

\begin{document}
\maketitle
\begin{abstract}
Existing LLM-enabled multi-agent frameworks are predominantly limited to digital or simulated environments and confined to narrowly focused knowledge domain, constraining their applicability to complex engineering tasks that require the design of physical embodiment, cross-disciplinary integration, and constraint-aware reasoning. This work proposes a multi-agent autonomous mechatronics design framework, integrating expertise across mechanical design, optimization, electronics, and software engineering to autonomously generate functional prototypes with minimal direct human design input. Operating primarily through a language-driven workflow, the framework incorporates structured human feedback to ensure robust performance under real-world constraints. To validate its capabilities, the framework is applied to a real-world challenge involving autonomous water-quality monitoring and sampling, where traditional methods are labor-intensive and ecologically disruptive. Leveraging the proposed system, a fully functional autonomous vessel was developed with optimized propulsion, cost-effective electronics, and advanced control. The design process was carried out by specialized agents, including a high-level planning agent responsible for problem abstraction and dedicated agents for structural, electronics, control, and software development. This approach demonstrates the potential of LLM-based multi-agent systems to automate real-world engineering workflows and reduce reliance on extensive domain expertise.
\end{abstract}    
\section{Introduction}
\label{sec:intro}

\begin{figure*} [tb]
  \centering
  \includegraphics[width=0.85\linewidth]{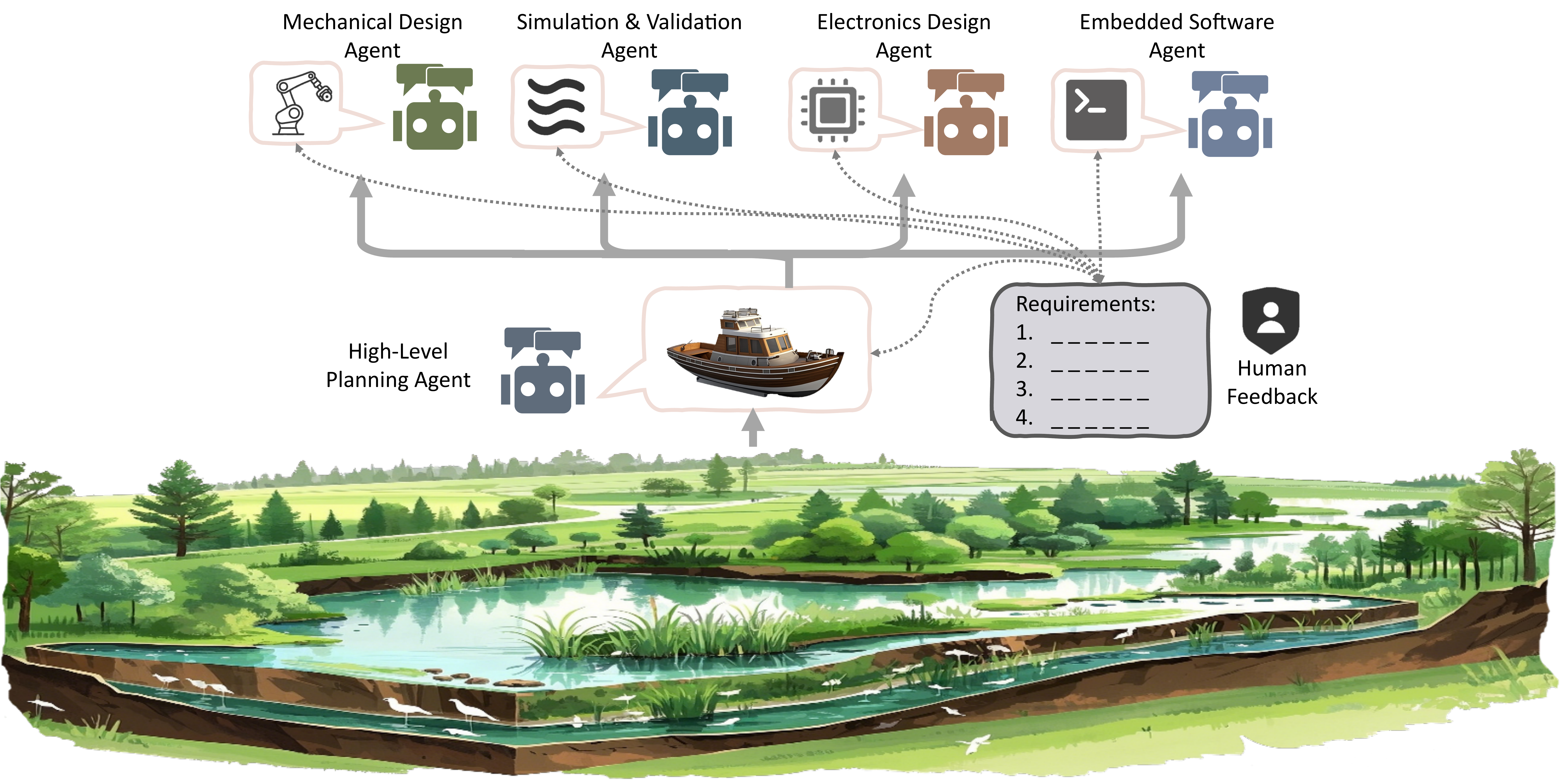}

   \caption{Conceptual architecture of the autonomous mechatronics design framework, illustrating multi-agent collaboration for system development with human-in-the-loop feedback.}
   \label{fig:01}
\end{figure*}

In the last few years, large language models (LLMs) \cite{changyupengSurveyEvaluationLarge2024,zhaoSurveyLargeLanguage2023,demszkyUsingLargeLanguage2023} have made rapid progress across a wide range of cognitive tasks, including natural language understanding \cite{minRecentAdvancesNatural2023}, code generation \cite{liuYourCodeGenerated2023,UsingLLMHelp}, symbolic reasoning \cite{mirzadehGSMSymbolicUnderstandingLimitations2024,panLogicLMEmpoweringLarge2023}, and scientific problem-solving \cite{wangSciBenchEvaluatingCollegeLevel2024,zhangScientificLargeLanguage2025}. Powered by transformer architectures \cite{khanTransformersVisionSurvey2022,vaswaniAttentionAllYou2017} and trained on massive dataset, models such as GPT-4 \cite{openaiGPT4TechnicalReport2024}, Claude \cite{priyanshuAIGovernanceAccountability2024}, DeepSeek \cite{deepseek-aiDeepSeekV3TechnicalReport2025}, and PaLM \cite{anilPaLM2Technical2023} exhibit strong performance in chain-of-thought reasoning \cite{weiChainofThoughtPromptingElicits2022,zhangChainPreferenceOptimization2024}, few-shot learning \cite{alayracFlamingoVisualLanguage2022,brownLanguageModelsAre2020}, and even multimodal understanding when extended to vision-language settings \cite{zhuMiniGPT4EnhancingVisionLanguage2023,kohGeneratingImagesMultimodal2023}. These capabilities have enabled LLMs to perform not only linguistic tasks, but also to engage in procedural synthesis \cite{liGuidingEnumerativeProgram2024}, and structured decision-making \cite{liPreTrainedLanguageModels2022,hagerEvaluationMitigationLimitations2024}, laying the foundation for their integration into agent-based systems capable of autonomous planning and tool use.

The rise of agent-based systems \cite{dinvernoUnderstandingAgentSystems2004} marks a critical milestone in artificial intelligence, enabling entities to autonomously perceive, reason, and act within specific environments \cite{beerDynamicalSystemsPerspective1995}. LLM-driven agents \cite{qiu2024llm} further enhance these capabilities through sophisticated linguistic comprehension and generation, proving effective in diverse applications such as text summarization \cite{vanveenAdaptedLargeLanguage2024}, software debugging \cite{tianDebugBenchEvaluatingDebugging2024,yangCOASTEnhancingCode2025}, documentation automation \cite{yeMPLUGDocOwlModularizedMultimodal2023,zhangChainAgentsLarge2024}, customer support \cite{karmakersantuGenerativeFeatureLanguage2016,peddintiUtilizingLargeLanguage2023}, mathematical theorem synthesis \cite{zhangMathemythsLeveragingLarge2024,ahnLargeLanguageModels2024}, virtual environment navigation \cite{zhouNavGPTExplicitReasoning2024}, and structured data querying \cite{panKwaiAgentsGeneralizedInformationseeking2024,zhuLargeLanguageModels2024}. In industry, LLM agents have streamlined narrowly defined workflows, including report generation \cite{wangRecMindLargeLanguage2024,zengEnhancingLLMsImpression2024,hanXBRLAgentLeveraging2024} and basic data analytics \cite{yangFinRobotOpenSourceAI2024,chughIntelligentAgentsDriven2023}, significantly improving operational efficiency.

However, current LLM-based agent implementations remain primarily confined to digital or simulated environments, thus limiting their practical application in complex engineering tasks which require the design of physical embodiment, cross-domain integration, and constraint-aware reasoning. Recent attempts to bridge this gap have emerged, demonstrating initial integration of LLM agents with physical experimentation in domains such as autonomous chemical synthesis \cite{m.branAugmentingLargeLanguage2024}, materials design \cite{ghafarollahiProtAgentsProteinDiscovery2024,ansariAgentbasedLearningMaterials2024,caldasramosReviewLargeLanguage2025}, drug discovery \cite{inoueDrugAgentExplainableDrug2024}, and adaptive multi-agent manufacturing systems \cite{xiaAutonomousSystemFlexible2023}. Despite these advancements, little attention has been given to LLM-enabled multi-agent frameworks targeting autonomous mechatronics design, a field inherently requiring multidisciplinary expertise across mechanical engineering, electronics, control systems, and software development. The inherent complexity of mechatronics design presents significant barriers to expertise and innovation, yet the cross-domain knowledge synthesis capabilities of LLMs offer emerging potential to support and accelerate certain aspects of mechatronics development. Leveraging LLMs could democratize access to mechatronics design, streamline the development process, and foster innovation by reducing the dependency on domain experts.
Addressing this research gap, we propose an LLM-enabled multi-agent framework for autonomous mechatronics system design, specifically targeting complex, interdisciplinary engineering challenges involving real-world physical embodiment. To illustrate the framework effectiveness, we apply it to an autonomous water-quality monitoring applications for wetlands \cite{ahmedWaterQualityMonitoring2019,gallantChallengesRemoteMonitoring2015}—an ecologically critical yet traditionally labor-intensive and expensive task. Leveraging specialized collaborative agents guided by structured human feedback, our approach significantly reduces complexity and streamline the processes. The resulting autonomous vessel demonstrates practical environmental monitoring potential, validating the proposed method and underscoring the potential of LLM-based agents in reshaping mechatronics design paradigms. This scalable framework could broadly facilitate future advances in areas such as biorobotics for precision medicine \cite{wang2023tactile,wang2024ai}, autonomous surgical robotics \cite{zhang2023step,chen2022path}, and AI-enabled multimodal health monitoring \cite{lo2024dietary}, significantly lowering barriers and expanding participation in interdisciplinary engineering innovation.

\section{Intelligence Levels Definition}
We proposed Intelligence Levels for Autonomous Mechatronics Design (AMD Levels) as followings:

\begin{itemize}
  \item \textbf{AMD Level 0 – Manual Design:} Entirely human-driven; all tasks and integration are performed manually.

  \item \textbf{AMD Level 1 – Assisted Automation:} Limited AI support (e.g., retrieval, basic code); decision-making remains fully human-controlled.

  \item \textbf{AMD Level 2 – Semi-Autonomous Design:} The system autonomously executes domain-specific tasks; humans manage integration and validation.

  \item \textbf{AMD Level 3 – Highly Autonomous Design:} Cross-domain coordination is automated; human involvement is restricted to high-level supervision and final evaluation.

  \item \textbf{AMD Level 4 – Fully Autonomous Design:} End-to-end automation. The system independently handles planning, integration, optimization, and validation, with human input limited to objectives setting and outcome assessment.
\end{itemize}

\section{Methodology}
\subsection{Framework design}

The proposed employs a hierarchical architecture governed by a High-Level Planning Agent, which decomposes complex challenges into modular tasks for domain-specific agents. Each agent autonomously generates, refines, and validates designs within its specialized discipline, while structured human input ensures feasibility and compliance with practical constraints.

Taking the autonomous water-quality monitoring and sampling system as an example, the High-Level Planning Agent initially analyzes functional requirements—such as mobility, environmental adaptability, and energy efficiency—to define the optimal configuration: a compact, dual-propeller vessel. Subsequently, the Mechanical Design Agent optimizes the propeller and hull structures, validated by the Simulation \& Validation Agent using computational fluid dynamics (CFD) and Finite Elements Analysis (FEA) for hydrodynamics and structural integrity. The Electronics Design Agent synthesizes propulsion and control circuits, leveraging available hardware to simplify integration. The Embedded Software Agent then generates motor-control and communication firmware, for optimal control of the hardware components. Meanwhile, human feedback is introduced as an external input, guiding the agents through ambiguous design choices and refining their outputs where limitations arise. This multi-agent approach integrates high-level objectives with low-level implementation, significantly reduce the iterative design process and dependency on domain experts, providing a scalable methodology for autonomous mechatronics design.

\subsection{High-Level Planning Agent}

The High-Level Planning Agent translates system-level requirements into actionable design objectives, guiding downstream engineering tasks. Leveraging LLM-based decision-making, it integrates functional goals, constraints, and human feedback into a unified design strategy. Mathematically, the agent’s function can be expressed as:
\begin{equation}
P = f(\bm{F}, \bm{C}, \bm{H})
\end{equation}
where $\bm{F}$ denotes the functional requirements (e.g., mobility, energy efficiency, environmental adaptability), $\bm{C}$ represents system constraints, and $\bm{H}$ refers to human feedback. By incorporating human feedback \cite{ouyangTrainingLanguageModels2022} directly into this loop, the agent ensures real-world insights continuously shape the design path, enabling robust, adaptable, and manufacturable mechatronic solutions.

\subsection{Mechanical Design Agent}
Mechanical design integrates structural mechanics \cite{hjelmstadFundamentalsStructuralMechanics2007}, fluid dynamics \cite{batchelorIntroductionFluidDynamics2000}, and manufacturability considerations \cite{guptaAutomatedManufacturabilityAnalysis1997} to ensure functional performance and production feasibility. Traditionally, engineers employ parametric modeling \cite{bolzonEffectiveComputationalTool2011}, finite element analysis \cite{kimIntroductionFiniteElement2018}, and iterative refinement, balancing constraints, such as material properties, hydrodynamic efficiency \cite{chandrasekharHydrodynamicHydromagneticStability2013}, and structural integrity. This requires substantial computational expertise, geometric reasoning, and practical fabrication experiences and knowledge. However, current LLMs face intrinsic limitations in mechanical design due to insufficient spatial reasoning and geometric cognition \cite{yamadaEvaluatingSpatialUnderstanding2024}. Unlike natural language processing or code generation, mechanical design inherently relies on real-time spatial visualization and parametric interaction—capabilities central to traditional CAD tools (e.g., SolidWorks \cite{SOLIDWORKS2023}, Fusion 360 \cite{AutodeskFusion3D}, COMSOL \cite{COMSOLMultiphysicsSoftware}) yet largely absent in existing LLMs. Additionally, mechanical design demands intuitive understanding of manufacturability constraints and material properties, which LLMs struggle to infer from text-based training alone.

To address these challenges, we propose a hybrid approach combining LLM-driven geometry generation and targeted human-in-the-loop feedback. This structured methodology forms the basis of our Mechanical Design Agent, enabling autonomous, and iterative optimization of mechanical structures yielding functional and effective mechanical designs. Specifically, the Mechanical Design Agent autonomously interprets high-level functional requirements to generate and refine mechanical components. For example, in designing the autonomous water-quality monitoring system, the agent iteratively optimizes the propeller and hull structures, systematically improving their stability, hydrodynamic efficiency, and manufacturability. This design workflow involves three structured stages: parametric geometric modeling, airfoil section generation, and transformation of 2D airfoil profiles into finalized 3D blade geometries.

Specifically, an LLM-driven mapping function formalizes this process:
\begin{equation}
\mathcal{L} \colon \mathcal{P} \to \mathcal{C}
\end{equation}
where $\mathcal{P}$ denotes the set of design parameters (e.g., blade root and tip chords, pitch angles, airfoil thickness, rake/skew angles) and $\mathcal{C}$ represents the generated CAD code. Iterative refinement is achieved via:
\begin{equation}
\mathcal{P}_{n+1} = \mathcal{F}(\mathcal{P}_n, R_n)
\end{equation}
with $R_n$ denoting simulation feedback at iteration $n$.

\noindent\textbf{a) Parametric Geometric Modeling}

The blade is defined along its span $z \in [0, L]$ by introducing the normalized coordinate:
\begin{equation}
r = \frac{z}{L}, \quad r \in [0, 1]
\end{equation}

Two primary functions govern the blade geometry:
Chord Distribution $C(z)$ is defined as:
\begin{equation}
C(z) = C_r + (C_t - C_r)r
\end{equation}
where $C_r$ and $C_t$ are the chord lengths at the root and tip, respectively. More complex variations (e.g., incorporating a Gaussian bulge) are expressed as:
\begin{equation}
C(z) = [C_r + (C_t - C_r)r] \left[1 + \beta \exp\left(-\gamma (r - r_0)^2\right)\right]
\end{equation}
with $\beta$, $\gamma$, and $r_0$ as design parameters.
Pitch Distribution $\alpha(z)$ is defined as:
\begin{equation}
\alpha(z) = \alpha_r + (\alpha_t - \alpha_r)r
\end{equation}
where $\alpha_r$ and $\alpha_t$ are the pitch angles (in radians) at the root and tip.

\vspace{1em}
\noindent\textbf{b) Airfoil Generation}

The agent generates a 2D airfoil profile at each spanwise station using a parameterized thickness distribution function $f(x)$, defined over the chord-wise coordinate $x \in [0, 1]$. In general, the upper and lower surfaces are given by:
\begin{equation}
\{(x - x_0, +f(x))\} \quad \text{and} \quad \{(x - x_0, -f(x))\}
\end{equation}
\vspace{1em}
\noindent\textbf{c) Transformation of 2D Airfoil Sections into a 3D Blade}

Each generated 2D airfoil section is transformed into its 3D location by applying:

\vspace{0.5em}
\noindent\textbf{Scaling:}
The normalized airfoil coordinates $(x, y)$ are scaled by the local chord $C(z)$:
\begin{equation}
X = C(z)x, \quad Y = C(z)S(z)y
\end{equation}
where $S(z)$ is an optional thickness scaling factor.

\vspace{0.5em}
\noindent\textbf{Rotation (Twist):}
The scaled coordinates are rotated by the local pitch angle $\alpha(z)$:
\begin{equation}
X_r = X\cos\alpha(z) - Y\sin\alpha(z)
\end{equation}
\begin{equation}
Y_r = X\sin\alpha(z) + Y\cos\alpha(z)
\end{equation}

\vspace{0.5em}
\noindent\textbf{Translation (Rake and Skew):}
To capture planform effects, a translation is applied:
\begin{equation}
S_{\text{rake}}(z) = -z\tan(\theta_{\text{rake}})
\end{equation}
\begin{equation}
S_{\text{skew}}(z) = rL\tan(\theta_{\text{skew}})
\end{equation}
The final coordinates become:
\begin{equation}
X_f = X_r + S_{\text{skew}}(z)
\end{equation}
\begin{equation}
Y_f = Y_r + S_{\text{rake}}(z)
\end{equation}

\begin{figure*} [tb]
  \centering
  \includegraphics[width=0.9\linewidth]{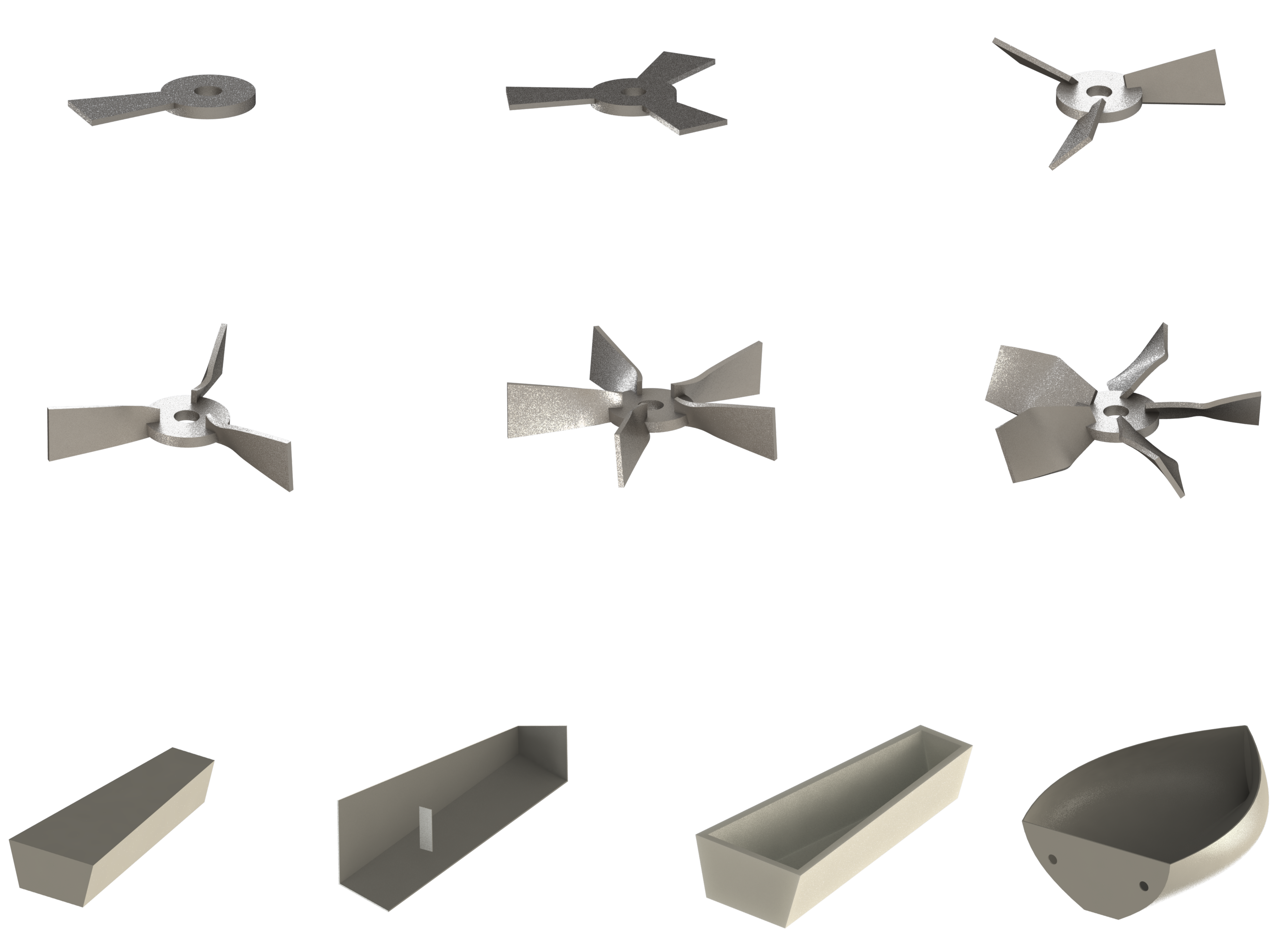}

   \caption{Iterative design process of propellers and hulls generated by the Mechanical Design Agent.}
   \label{fig:02}
\end{figure*}

\subsection{Simulation \& Validation Agent}
Structural validation is essential for mechatronic systems, ensuring robustness and reliability under operational conditions. Traditionally, engineers employ FEA and computational methods to assess structural integrity, deformation, and stress distribution, guiding material selection, geometric optimization, and load-bearing capacity. Integrating LLMs into simulation-driven workflows could accelerate the in structural validation tasks such as model configuration and boundary condition definition. However, current LLMs struggle with interpreting complex multiphysics phenomena, non-linear material behaviors, and ensuring numerical convergence, thus necessitating a hybrid approach that combines AI-driven methodologies with empirical validation.

Within the proposed framework, the Simulation \& Validation Agent autonomously validates mechanical structures generated by the Mechanical Design Agent. Specifically, it automatically configures simulations, defines materials and boundary conditions, and performs physics-based analyses using finite element methods. Results from these simulations guide iterative design refinement to enhance structural performance and manufacturability. For validating the autonomous boat’s propulsion system, the agent employs CFD and structural FEA, evaluating fluid-structure interactions, mechanical stresses under operational loads, and potential fatigue points.

Initially, to establish baseline safety margins, the agent performed a fully coupled fluid–structure interaction (FSI) simulation under laminar flow conditions as a baseline analysis, offering insights into stress distribution and potential failure regions under extreme passive exposure. Specifically, the agent sets up a FSI interface \cite{dowellMODELINGFLUIDSTRUCTUREINTERACTION2001} to transfer pressure and shear to the blades, followed by a stationary stress analysis computing von Mises stress:
\begin{equation}
\sigma_{vM} = \sqrt{0.5 \left[ (\sigma_1 - \sigma_2)^2 + (\sigma_2 - \sigma_3)^2 + (\sigma_3 - \sigma_1)^2 \right]}
\end{equation}
with $\sigma_1$, $\sigma_2$, and $\sigma_3$ being the principal stresses.

Subsequently, to model the propeller’s high-speed rotation and fluid interaction, the agent uses rotating machinery analysis in COMSOL, solving the Reynolds-Averaged Navier–Stokes (RANS) equations \cite{alfonsiReynoldsAveragedNavierStokes2009} with the Shear Stress Transport (SST) turbulence model \cite{menterReviewShearstressTransport2009}. A rotating reference frame is applied to the propeller domain, with the surrounding fluid kept stationary to ensure proper interface conditions. The momentum equation in the rotating frame is given by:
\begin{equation}
\rho (\mathbf{u} \cdot \nabla \mathbf{u}) = -\nabla p + \nabla \cdot \left[ \mu_{\text{eff}} (\nabla \mathbf{u} + (\nabla \mathbf{u})^T) \right] + F_{\text{rot}}
\end{equation}
where $\rho$ is the fluid density, $\mathbf{u}$ is the velocity field, $p$ is the pressure, and $\mu_{\text{eff}} = \mu_t + \mu$ is the effective viscosity accounting for both molecular ($\mu$) and turbulent ($\mu_t$) contributions. The term $F_{\text{rot}}$ encompasses Coriolis and centrifugal forces. By blending $k$--$\omega$ and $k$--$\epsilon$ formulations \cite{rahmanWalldistancefreeFormulationSST2019,parenteImprovedKeModel2011}, the SST model ensures accurate near-wall treatment and robust free-stream predictions. 



\subsection{Electronics Design Agent}

Electrical engineering design involves circuit analysis, electronics design, signal processing, control systems, and embedded computing to develop robust electronic systems. Traditional workflows include schematic design, component selection, PCB layout \cite{yanRecentResearchDevelopment2010}, and system-level testing, which require expertise in electromagnetic compatibility (EMC), thermal management, and manufacturability. Recent advancements in LLMs offer transformative potential by assisting in tasks, such as automated component selection \cite{taffernerComparingLargeLanguage2024}, schematic generation, and design optimization. Trained on extensive technical documentation, these models can efficiently parse datasheets, recommend suitable components, and streamline procurement processes, significantly reducing design cycles.

However, LLMs currently face limitations in real-time design constraints, and robustness under practical constraints. To overcome these limitations, the Electronics Design Agent is designed to employ a hybrid workflow that integrates LLM-based reasoning, structured human feedback, and existing hardware utilization. Rather than designing circuits from scratch, the agent first evaluates available hardware resources for potential reuse, optimizing resource efficiency and accelerating development.

\begin{figure*} [tb]
  \centering
  \includegraphics[width=1\linewidth]{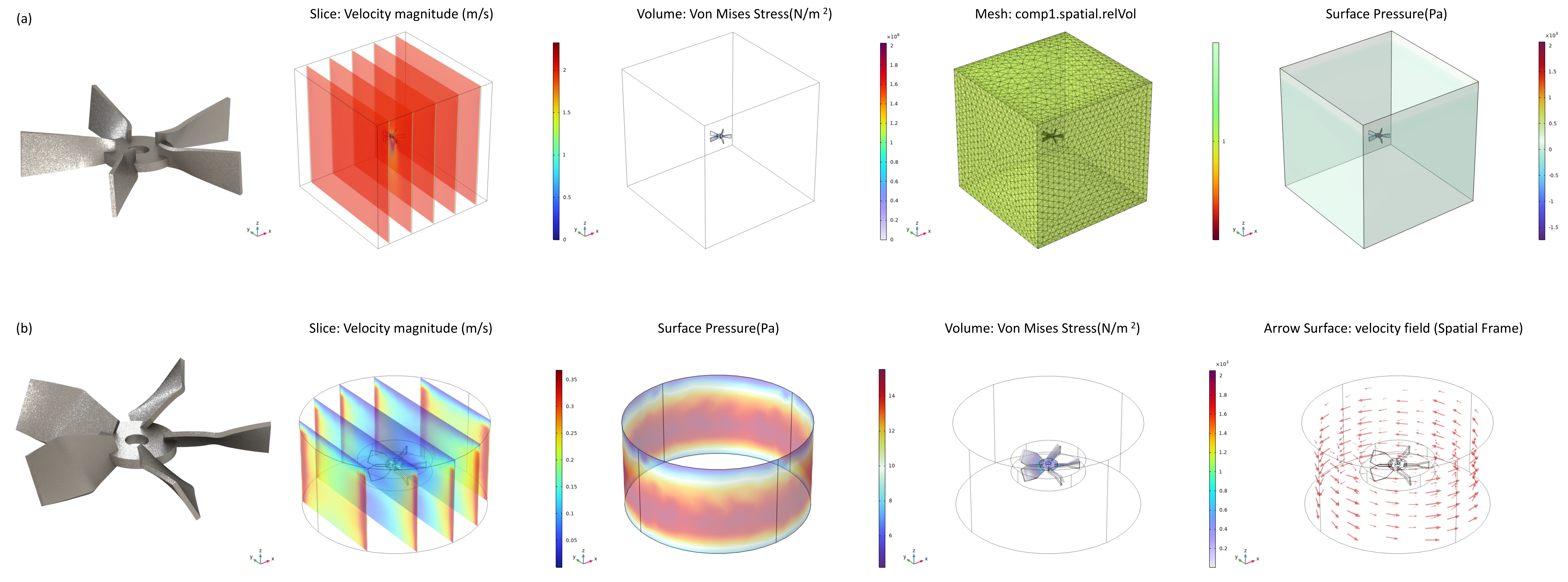}

   \caption{CFD and structural analysis of the optimized propeller design performed by the Simulation \& Validation Agent. (a) Structural stress distribution under laminar flow conditions (baseline analysis). (b) Results obtained from rotating turbulent flow analysis using the Shear Stress Transport (SST) model.Results demonstrate stable hydrodynamic performance, reasonable structural stress levels, and highlight areas for further refinement.}
   \label{fig:03}
\end{figure*}

\subsection{Embedded Software Agent}
Embedded systems serve as computational backbones in mechatronics design, translating high-level objectives into precise control commands \cite{heathEmbeddedSystemsDesign2002}. Effective embedded software design requires expertise in real-time control algorithms, communication protocols, firmware development, and sensor-driven actuation. Recent advances in LLMs offer opportunities for automating embedded software tasks, including firmware generation, adaptive control logic optimization, and efficient resource allocation. Trained on extensive programming datasets, LLMs can autonomously generate structured code and adapt existing control strategies. However, current LLM-driven code still faces limitations in real-time execution efficiency and direct peripheral interaction, thus necessitating empirical validation.

Within our framework, the Embedded Software Agent translates high-level specifications into executable firmware tailored specifically to hardware architectures defined by the Electronics Design Agent. This inter-agent coordination ensures seamless integration between hardware and software. Additionally, the agent implements adaptive motor-control strategies essential for tasks such as differential speed control in rudderless propulsion systems, dynamically adjusting motor actuation to achieve reach the targets precisely under varying operational conditions.

\section{Experimental Results}
\subsection{Mechanical Design Agent}
Guided by the autonomous boat objective established by the High-Level Planning Agent, the Mechanical Design Agent iteratively optimized the propeller and hull structures, as shown in ~\cref{fig:02}. The propeller design process, illustrated in the top two rows of ~\cref{fig:02}, began with a three-blade configuration featuring a blade length of 26 mm and a hub diameter of 20 mm. Early iterations exhibited poor hydrodynamic performance. Through successive refinements in blade curvature, twist angle, and hub-to-blade ratio, the agent improved hydrodynamic efficiency by enhancing the lift-to-drag ratio and mitigating flow separation. The final geometry balances structural rigidity with stable thrust generation.

Retrieving the detailed implementation of agent, the agent tailored chord and pitch distributions to maximize mid-span loading using piecewise interpolation, as shown below:
\begin{equation}
C(z) =
\begin{cases}
  C_r + (C_m - C_r)(2r),       & 0 \leq r \leq 0.5 \\
  C_m + (C_t - C_m)(2r + 1),   & 0.5 \leq r \leq 1
\end{cases}
\end{equation}

with $C_m$ as the maximum chord at mid-span. The pitch distribution is similarly interpolated:
\begin{equation}
\alpha(z) =
\begin{cases}
  \alpha_r + (\alpha_m - \alpha_r)(2r),       & 0 \leq r \leq 0.5 \\
  \alpha_m + (\alpha_t - \alpha_m)(2r + 1),   & 0.5 \leq r \leq 1
\end{cases}
\end{equation}

Generic airfoil sections were generated and transformed into a finalized three-dimensional blade geometry optimized for thrust.

The bottom row of ~\cref{fig:02} illustrates the hull design evolution. Initial outputs consisted of solid rectangular forms that failed to satisfy buoyancy and weight constraints. Later iterations introduced hollowed structures, though early designs lacked viable compartmentalization. Through progressive refinement, the agent incorporated internal reinforcements, strategic cutouts, and streamlined contours, resulting in a curved bow design that reduces drag, improves maneuverability, and accommodates onboard components. These results demonstrate the Mechanical Design Agent’s capacity for autonomous structural refinement under evolving constraints, producing mechanically feasible and performance-optimized components suitable for downstream validation

\begin{figure*} [tb]
  \centering
  \includegraphics[width=0.8\linewidth]{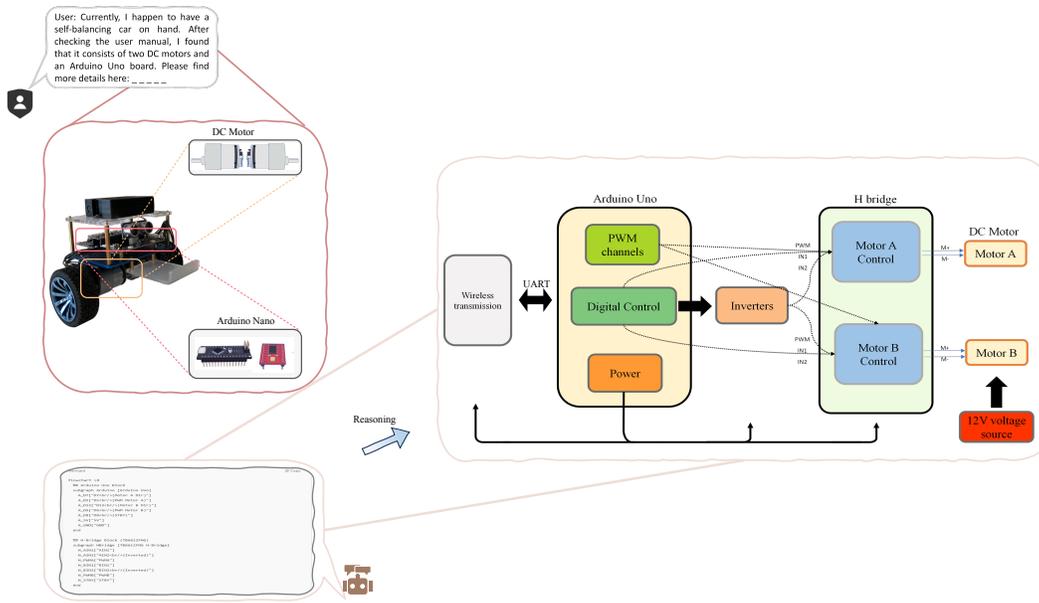}

   \caption{ Electronics Design Agent workflow based on user feedback. The agent analyzes the existing balance car system and reconstructs the core control architecture, including Arduino-based PWM control and H-bridge motor driving}
   \label{fig:04}
\end{figure*}

\subsection{Simulation \& Validation Agent }

To evaluate the propeller design generated by the Mechanical Design Agent, the Simulation \& Validation Agent conducted a two-stage CFD analysis, as shown in ~\cref{fig:03}. For the initial baseline analysis, the agent autonomously performed a laminar-flow fluid–structure interaction simulation using, as an example, the fifth design iteration shown in ~\cref{fig:02}. The resulting velocity magnitude slices and surface pressure distributions confirmed smooth flow behavior, minimal separation, and coherent pressure fields, validating early-stage design stability. Given the high Reynolds number regime of the final propeller configuration, the agent transitioned to a turbulence-resolving analysis, employing the SST model for enhanced fidelity in near-wall flow characterization, again exemplified by the final iteration shown in ~\cref{fig:02}. Unlike the baseline scenario, however, the agent could not autonomously configure this turbulent-flow analysis, requiring substantial human feedback and expert assistance to iteratively adjust solver settings, boundary conditions, and spatial discretization until convergence. Further detailed validation of these results remains necessary to fully ensure accuracy and reliability.

Despite demonstrating physics module inference, the Simulation \& Validation Agent remains limited in execution. Configuration inconsistencies—partly due to software interface variability—undermine its reliability across solver settings. Defining boundary conditions and spatial domain partitioning remains particularly challenging, constrained by the agent’s lack of geometric intuition and visuospatial inference capability. These limitations are pronounced in transient, multiphysics simulations, where complex temporal and inter-domain dependencies often lead to convergence failure. Addressing these challenges will require integration of structured expert priors, few-shot learning from validated workflows, and physics-informed guidance to systematically elevate robustness and generalizability.

\begin{figure*} [tb]
  \centering
  \includegraphics[width=1\linewidth]{figures/05.pdf}

   \caption{Embedded control logic and human-verified signal outputs. The agent generates control logic based on the Electronics Design Agent’s architecture, while a logic analyzer is used to capture and verify PWM signals under different scenarios.}
   \label{fig:05}
\end{figure*}

\subsection{Electronics Design Agent}
To evaluate the agent’s adaptive design capability, an Arduino Nano-based self-balancing car was introduced as external input. The agent received specifications of the system, including motor drivers and power modules, and assessed whether its components could be disassembled and reconfigured into a functional propulsion system for the autonomous boat.

As shown in ~\cref{fig:04}, the Electronics Design Agent synthesized a system architecture that repurposes the self-balancing car hardware for dual-motor actuation. Leveraging circuit knowledge and structured human input, the agent analyzed the feasibility of reusing the motor control, power distribution, and signal processing elements to construct an efficient and compact drive system.

The resulting architecture comprises two subsystems: the Arduino Nano control unit and an H-bridge motor driver. Pulse-width modulation (PWM) signals from the control unit are processed via logic circuitry and inverters to regulate motor speed and direction. The H-bridge enables bidirectional current flow for two independently controlled DC motors, each powered by a 12V source. Differential motor speed control provides rudderless steering, enabling smooth turning and precise navigation—crucial for autonomous vessels operations. Wireless transmission further supports remote operation in real-world environments.

\subsection{Embedded Software Agent}
\cref{fig:05} presents the embedded firmware architecture structured by the agent. The system initializes by configuring GPIOs and peripherals, then continuously monitors incoming serial data. Upon receiving a complete command, it parses the motion type—such as forward, backward, left, right, or stop—and generates corresponding PWM signals to control Motor A and Motor B via an H-bridge driver. For turning maneuvers, the agent applies differential velocity control, adjusting motor speeds to modulate turning radius without the need for a physical rudder.

Real-world execution is assessed using logic analyzer traces, as shown in Fig. 5, which capture IN1, IN2, and PWM signals for both motors under various commands. The observed duty cycles—39.2\% (forward), 31.4\% (backward), 58.8\% (left), and 15.7\% (right)—confirm precise motor actuation aligned with the embedded control logic. Directional transitions are accurately executed through appropriate phase shifting. A minor deviation was observed in the first PWM pulse during turning, attributed to initialization latency from transient register updates. This effect is brief and would not affect the system-level control performance, affirming the agent's ability to autonomously generate robust, real-time motor control code suitable for autonomous navigation.

\section{Discussion}
The proposed framework is designed to assist the design of functioning and effective mechatronic systems with structured human feedback. Human guidance proved crucial, particularly for defining modeling sequences in mechanical design tasks. Embedded Software agents demonstrated higher autonomy given clear architectural specifications provided by the Electronics agents, while simple simulations can be autonomously executed via the COMSOL–MATLAB interface, more complex setups—especially involving domain definition and intricate boundary conditions in multiphysics modules such as transient fluid simulations—remain highly challenging, suggesting the potential benefits of few-shot learning and structured prior injection. Future work will address hardware optimization, such as waterproofing electronic components, and leverage multimodal LLM capabilities, particularly vision-based functionalities, to enhance autonomous control logic.

\section{Conclusion}
This work introduces an LLM-enabled multi-agent framework that advances autonomous mechatronics design across disciplines boundaries. By integrating language-driven intelligence with structured human feedback, it redefines engineering workflows, lowers expertise barriers, and signals a broader shift toward scalable, democratized innovation in physical-system design.

{
    \small
    \bibliographystyle{unsrtnat}
    \bibliography{main}
}


\end{document}